\documentclass[11pt,a4paper]{article}
\usepackage{authblk}
\usepackage[hyperref]{acl2020}
\usepackage{times}
\usepackage{latexsym}

\usepackage{microtype}

\aclfinalcopy 

\usepackage{multirow}
\usepackage{multicol}
\usepackage{latexsym,amsmath}
\usepackage{graphicx}
\newcommand\wenc{\textsc{WordEnc}\ }
\newcommand\senc{\textsc{SentEnc}\ }

\newcommand\hpe{\textsc{HPE}\ }
\newcommand\vX{\pmb{X}}

\newcommand\vz{\pmb{z}}
\newcommand\vy{\pmb{y}}

\newcommand\bos{$<\textsc{BOS}>$\ }
\newcommand\eos{$<\textsc{EOS}>$\ }

\newcommand{\seq}{Seq2Seq + att + AE}
\newcommand{\tseq}{TransSeq2Seq + AE}
\newcommand{\hseq}{HierSeq2Seq + AE }
\newcommand{\htseq}{HierTransSeq2Seq + AE}

\title{Question Generation from Paragraphs: A Tale of Two Hierarchical Models}
\author[1,2,3]{Vishwajeet Kumar}
\author[2]{Raktim Chaki}
\author[2]{Sai Teja Talluri}
\author[2]{Ganesh Ramakrishnan}
\author[3]{Yuan-Fang Li}
\author[3]{Gholamreza Haffari}
\affil[1]{IITB-Monash Research Academy, Mumbai India}
\affil[2]{IIT Bombay, Mumbai, India}
\affil[3]{Monash University, Australia}
\affil[ ]{\texttt{\{vishwajeet, ganesh\}@cse.iitb.ac.in}}
\affil[ ]{\texttt{\{rcraktimc, saiteja.talluri\}@gmail.com}}
\affil[ ]{\texttt{\{yuanfang.li, gholamreza.haffari\}@monash.edu}}
\begin{document}
\maketitle
\begin{abstract}
Automatic question generation from paragraphs is an important and challenging problem, particularly due to the long context from paragraphs. In this paper, we propose and study two hierarchical models for the task of question generation from paragraphs. Specifically, we propose (a) a novel hierarchical BiLSTM model with selective attention and (b) a novel hierarchical Transformer architecture, both of which learn hierarchical representations of paragraphs. 
We model a paragraph in terms of its constituent sentences, and a sentence in terms of its constituent words. While the introduction of the attention mechanism benefits the hierarchical BiLSTM model, the hierarchical Transformer, with its inherent attention and positional encoding mechanisms also performs better than flat transformer model.
We conducted empirical evaluation on the widely used SQuAD and MS MARCO datasets using standard metrics. 
The results demonstrate the overall effectiveness of the hierarchical models over their flat counterparts. 
Qualitatively, our hierarchical models are able to generate fluent and relevant questions.
\end{abstract}


\section{Introduction}
Question Generation (QG) from text has gained significant popularity in recent years in both academia and industry, owing to its wide applicability in a range of scenarios including conversational agents, automating reading comprehension assessment, and improving question answering systems by generating additional training data.
Neural network based methods represent the state-of-the-art for automatic question generation. These models do not require templates or rules, and are able to generate fluent, high-quality questions. 

Most of the work in question generation takes sentences as input \citep{cardie2018harvesting,kumar2018automating,song2018leveraging,kumar2018framework}. QG at the paragraph level is much less explored and it has remained a challenging problem. The main challenges in paragraph-level QG stem from the larger context that the model needs to assimilate in order to generate relevant questions of high quality.

Existing question generation methods are typically based on recurrent neural networks (RNN), such as bi-directional LSTM. Equipped with different enhancements such as the attention, copy and coverage mechanisms, RNN-based models~\citep{du2017learning,kumar2018automating,song2018leveraging} achieve good results on sentence-level question generation. However, due to their ineffectiveness in dealing with long sequences, paragraph-level question generation remains a challenging problem for these models. 

Recently, \cite{zhao2018paragraph} proposed a paragraph-level QG model with maxout pointers and a gated self-attention encoder. To the best of our knowledge this is the only model that is designed to support paragraph-level QG and outperforms other models on the SQuAD dataset~\citep{rajpurkar2016squad}. One straightforward extension to such a model would be to reflect the structure of a paragraph in the design of the encoder. Our first attempt is indeed a hierarchical BiLSTM-based paragraph encoder ( \hpe), wherein, the hierarchy comprises the word-level encoder that feeds its encoding to the sentence-level encoder.  Further, dynamic paragraph-level contextual information in the BiLSTM-\hpe is incorporated via both word- and sentence-level selective attention.

However, LSTM is based on the recurrent architecture of RNNs, making the model somewhat rigid and less dynamically sensitive to different parts of the given sequence. Also LSTM models are slower to train. In our case, a paragraph is a sequence of sentences and a sentence is a sequence of words. The Transformer~\citep{vaswani2017attention} is a recently proposed neural architecture designed to address some deficiencies of RNNs. Specifically, the Transformer is based on the (multi-head) attention mechanism, completely discarding recurrence in RNNs. This design choice allows the Transformer to effectively attend to different parts of a given sequence. Also Transformer is relatively much faster to train and test than RNNs. 

As humans, when reading a paragraph, we look for important sentences first and then important keywords in those sentences to find a concept around which a question can be generated. Taking this inspiration, we give the same power to our model by incorporating word-level and sentence-level selective attention to generate high-quality questions from paragraphs. 

In this paper, we present and contrast novel approachs to QG at the level of paragraphs. Our main contributions are as follows:

\begin{itemize}
    \item  We present two hierarchical models for encoding the paragraph based on its structure. We analyse the effectiveness of these models for the task of automatic question generation from paragraph.

    \item Specifically, we propose a novel hierarchical Transformer architecture. At the lower level, the encoder first encodes words and produces a sentence-level representation. At the higher level, the encoder aggregates the sentence-level representations and learns a paragraph-level representation. 
    
    \item We also propose a novel hierarchical BiLSTM model with selective attention, which learns to attend to important sentences and words from the paragraph that are relevant to generate meaningful and fluent questions about the encoded answer.

    \item We also present attention mechanisms for dynamically incorporating contextual information in the hierarchical paragraph encoders and experimentally validate their effectiveness.
    

\end{itemize}

\section{Related Work}

Question generation (QG) has recently attracted significant interests in the natural language processing (NLP)~\citep{du2017learning,kumar2018automating,song2018leveraging,kumar2018framework} and computer vision (CV)~\citep{li2018visual,fan2018question} communities. 
Given an input ({\em e.g.},\ a passage of text in NLP or an image in CV), optionally also an answer, the task of QG is to generate a natural-language question that is answerable from the input. 

Existing text-based QG methods can be broadly classified into three categories: (a) rule-based methods, (b) template-base methods, and (c) neural network-based methods. Rule based methods~\citep{heilman2010good} perform syntactic and semantic analysis of sentences and apply fixed sets of rules to generate questions. They mostly rely on syntactic rules written by humans~\citep{heilman2011automatic} and these rules change from domain to domain. On the other hand, template based methods~\citep{ali2010automation} use generic templates/slot fillers to generate questions. More recently, neural network-based QG methods~\citep{du2017learning,kumar2018automating,song2018leveraging} have been proposed. They employ an RNN-based encoder-decoder architecture and train in an end-to-end fashion, without the need of manually created rules or templates. 

\cite{du2017learning} were the first to propose a sequence-to-sequence (Seq2Seq) architecture for QG.  \cite{kumar2018automating} proposed to augment each word with linguistic features and encode the most relevant \emph{pivotal answer} in the text while generating questions. Similarly, \cite{song2018leveraging} encode ground-truth answers (given in the training data), use the copy mechanism and additionally employ context matching to capture interactions between the answer and its context within the passage. They encode ground-truth answer for generating questions which might not be available for the test set. 

\cite{zhao2018paragraph} recently proposed a Seq2Seq model for paragraph-level question generation, where they employ a maxout pointer mechanism with a gated self-attention encoder. 
\cite{tran2018importance} contrast recurrent and non-recurrent architectures on their effectiveness in capturing the hierarchical structure. In Machine Translation, non-recurrent model such as a Transformer~\citep{vaswani2017attention} that does not use convolution or recurrent connection is often expected to perform better. However, Transformer, as a non-recurrent model, can be more effective than the recurrent model because it has full access to the sequence history. Our findings also suggest that LSTM outperforms the Transformer in capturing the hierarchical structure. In contrast, \cite{goldberg2019assessing} report settings in which attention-based models, such as BERT are better capable of learning hierarchical structure than LSTM-based models.

\section{Hierarchical Paragraph Representation}
We propose a general hierarchical architecture for better paragraph representation at the level of words and sentences. This architecture is agnostic to the type of encoder, so we base our hierarchical architectures on BiLSTM and Transformers. We then present two  decoders (LSTM and Transformer) with hierarchical attention over the paragraph representation, in order to provide  the dynamic context needed by the decoder. The decoder is further conditioned on the provided (candidate) answer to generate relevant questions. 

\paragraph{Notation:} The question generation task consists of pairs $(\vX,\pmb{y})$ conditioned on an encoded answer $\vz$, where $\vX$ is a paragraph, and $\vy$ is the target question which needs to be generated with respect to the paragraph.. 
Let us denote the $i$-th sentence in the paragraph by $\pmb{x}_i$, where $x_{i,j}$ denotes the $j$-th word of the sentence. 
We assume that the first and last words of the sentence are special beginning-of-the-sentence $<\textsc{BOS}>$ and end-of-the-sentence $<\textsc{EOS}>$ tokens, respectively.

\subsection{Hierarchical Paragraph Encoder}
\label{sec:hpe}

Our hierarchical paragraph encoder (\hpe) consists of two encoders, {\em viz.}, a sentence-level and a word-level encoder; ({\em c.f.} Figure~\ref{fig:fig1}). 

\paragraph{Word-Level Encoder:} The lower-level encoder \wenc encodes the words of individual sentences. This encoder produces a sentence-dependent word representation $\pmb{r}_{i,j}$ for each word $x_{i,j}$ in a sentence $\pmb{x}_i$, {\em i.e.},  $\pmb{r}_i=\wenc(\pmb{x}_i)$. 
This representation is the output of the last encoder block in the case of Transformer, and the last hidden state in the case of BiLSTM.
Furthermore, we can produce a fixed-dimensional representation for a sentence as a function of $\pmb{r_i}$, {\em e.g.}, by summing (or averaging) its contextual word representations, or 
concatenating the contextual representations of its \bos  and \eos tokens. 
We denote the resulting sentence representation by $\tilde{\pmb{s}}_i$ for a sentence $\pmb{x}_i$.

\paragraph{Sentence-Level Encoder:} At the higher level, our HPE consists of another  encoder to produce paragraph-dependent representation for the sentences. 
The input to this encoder are the sentence representations produced by the lower level encoder, which are insensitive to the paragraph context. In the case of the transformer, the sentence representation is combined with its positional embedding to take the ordering of the paragraph sentences into account.
The output of the higher-level encoder is contextual representation for each set of sentences $\pmb{s}=\senc(\tilde{\pmb{s}})$, where $\pmb{s}_i$ is the paragraph-dependent representation for the $i$-th sentence. 

In the following two sub-sections, we present our two  hierarchical encoding architectures, {\em viz.}, the hierarchical BiLSTM in Section \ref{sec:hbilstm}) and hierarchical transformer in Section \ref{sec:htrans}). 

\subsection{Dynamic Context in BiLSTM-\hpe}
\label{sec:hbilstm}

\begin{figure*}[htb]
    \centering
    \includegraphics[width=\textwidth]{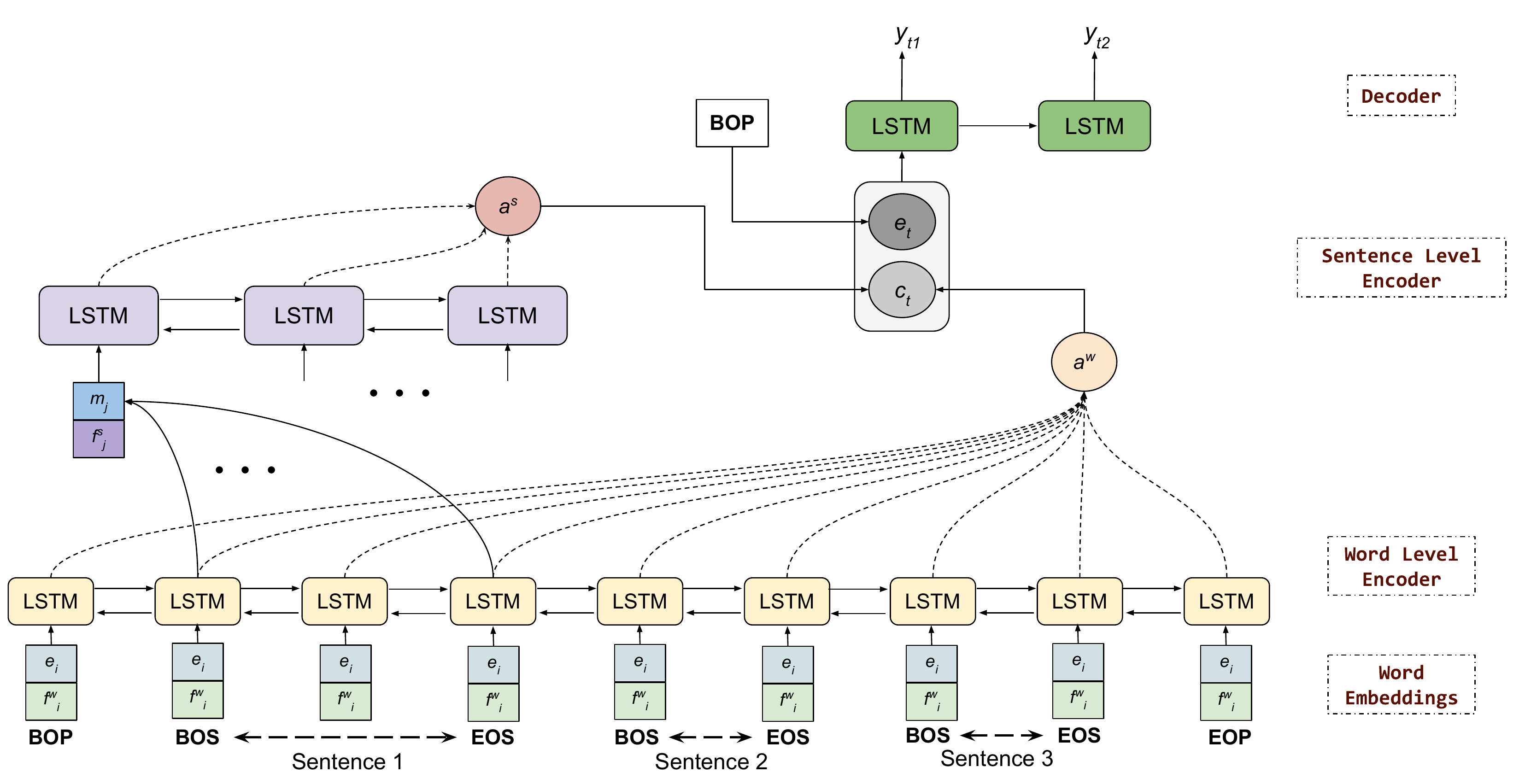}
    \caption{Our hierarchial LSTM Architecture.}
    \label{fig:fig1}
\end{figure*}

In this first option, {\em c.f.}, Figure \ref{fig:fig1}, we use both word-level attention and sentence level attention  in a Hierarchical BiLSTM encoder to obtain the hierarchical paragraph representation. We employ the attention mechanism proposed in \citep{Luong2015EffectiveAT} at both the word and sentence levels. We employ the BiLSTM (Bidirectional LSTM) as both, the word as well as the sentence level encoders. We concatenate forward and backward hidden states to obtain sentence/paragraph representations. Subsequently, we employ a unidirectional LSTM unit as our decoder, that generates the target question one word at a time, conditioned on (i) all the words generated in the previous time steps and (ii) on the encoded answer. The methodology employed in these modules has been described next.

\paragraph{Word-level Attention:} We use the LSTM decoder's previous hidden state and the word encoder's hidden state to compute attention over words (Figure~\ref{fig:fig1}). We the concatenate forward and backward hidden states of the BiLSTM encoder to obtain the final hidden state 
representation ($h_t$) at time step t. Representation ($h_t$) is calculated as:
$\pmb{h}_{t} =  \wenc(\mathbf{h}_{t-1}, [\pmb{e}_{t}, \pmb{f}^{w}_{t}])$,  where $\pmb{e}_{t}$ represents the GLoVE \citep{pennington-etal-2014-glove} embedded representation of word ($x_{i,j}$) at time step t and  $\pmb{f}^{w}_{t}$ is the embedded BIO feature for answer encoding.\\
The word level attention ($\pmb{a}^{w}_{t}$) is computed as:  $\pmb{a}^{w}_{t} = Softmax({[{u}^{w}_{t_i}]}^{M}_{i=1})$, where M is the number of words, and $ u^{w}_{t_i} = \pmb{v}^{T}_{w} \tanh(\mathbf{W}_{w} [\pmb{h}_{i},\pmb{d}_{t}])$ and $\pmb{d}_{t}$ is the decoder hidden state at time step t.

\par We calculate sentence representation ($\tilde{\pmb{s}}_i$) using word level encoder's hidden states as: $\tilde{\pmb{s}}_i= \frac{1}{|\pmb{x}_{i}|} \sum_{j} \pmb{r}_{i,j}$, where $\pmb{r}_{i,j}$ is the word encoder hidden state representation of the $j^{th}$ word of the $i^{th}$ sentence.

\paragraph{Sentence-level Attention:} We feed the sentence representations $\tilde{\pmb{s}}$ to our sentence-level BiLSTM encoder ({\em c.f.} Figure~\ref{fig:fig1}). Similar to the word-level attention, we again the compute attention weight over every sentence in the input passage, using (i) the previous decoder hidden state and (ii) the sentence encoder's hidden state. As before, we concatenate the forward and backward hidden states of the sentence level encoder to obtain the final hidden state representation. The hidden state ($\pmb{g}_{t}$) of the sentence level encoder is computed as: $\pmb{g}_{t} =  \senc(\pmb{g}_{t-1}, [\tilde{\pmb{s}}_{t}, \pmb{f}^{s}_{t}])$, where $\pmb{f}^{s}_{t}$ is the embedded feature vector denoting whether the sentence contains the encoded answer or not.

\par The selective sentence level attention ($\pmb{a}^{s}_{t}$) is computed as: $\pmb{a}^{s}_{t} = Sparsemax({[{u}^{w}_{t_i}]}^{K}_{i=1})$, where, K is the number of sentences,  $u^{s}_{t_i} = \pmb{v}^{T}_{s} \tanh(\mathbf{W}_{s} [\pmb{g}_{i},\pmb{d}_{t}])$.

The final context ($\pmb{c}_t$) based on hierarchical selective attention is computed as:   $\pmb{c}_{t} = \sum_{i} {a}^{s}_{t_i} \sum_{j}\overline{a}^{w}_{t_{i,j}}    \pmb{r}_{i,j}$, where $\overline{a}^{w}_{t_{i,j}}$ is the word attention score obtained from $\pmb{a}^{w}_{t}$ corresponding to $j^{th}$ word of the $i^{th}$ sentence.

The context vector $\pmb{c}_t$ is fed to the decoder at time step t along with embedded representation of the previous output.

\subsection{Dynamic Context in Transformer-\hpe \label{sec:htrans}} 
\begin{figure}[htb]
    \centering
    \includegraphics[scale=0.35]{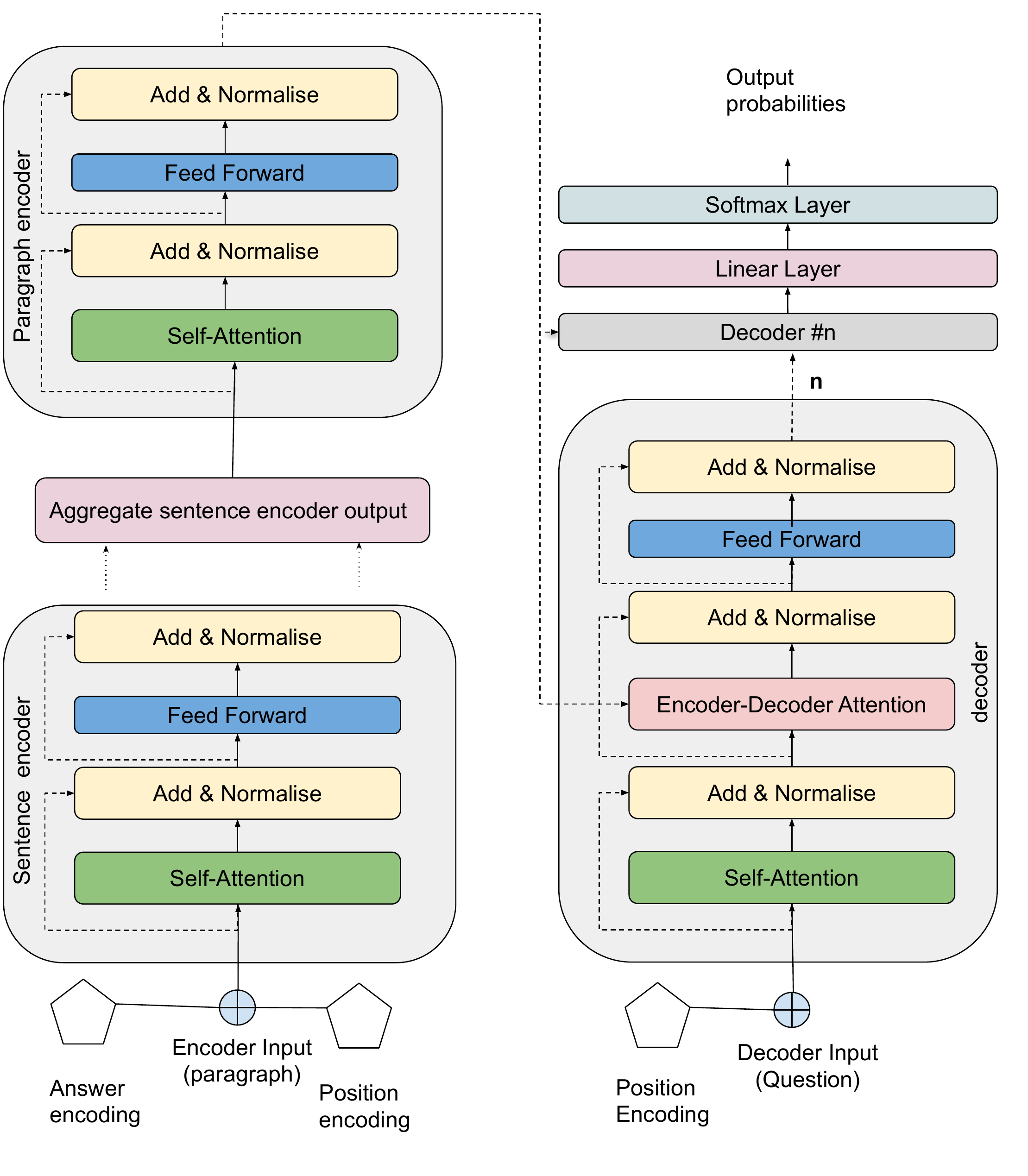}
    \caption{Our hierarchical Transformer architecture.}
    \label{fig:Hierarchical transformer}
\end{figure}

In this second option ({\em c.f.} Figure~\ref{fig:Hierarchical transformer}), we make use of a Transformer decoder to generate the target question, one token at a time, from left to right. 
For generating the next token, the decoder attends to the previously generated tokens in the question, the encoded answer and the paragraph. 
We postulate that attention to the paragraph benefits from our hierarchical representation, described in Section \ref{sec:hpe}.
That is, our model  identifies firstly the relevance of the sentences, and then the relevance of the words within the sentences. This results in a hierarchical attention module (HATT) and its multi-head extension (MHATT), which replace the attention mechanism to the source in the Transformer decoder. 
We first explain the sentence and paragragh encoders (Section~\ref{sec:sentparaenc}) before moving on to explanation of the decoder (Section~\ref{sec:decoder}) and the hierarchical attention modules (HATT and MHATT in Section~\ref{sec:mhatt}).

\subsubsection{Sentence and paragraph encoder \label{sec:sentparaenc}}
The sentence encoder transformer maps an input sequence of word representations $\textbf{x} = (x_0,\cdots,x_n)$ to a sequence of continuous sentence representations $\textbf{r} = (r_0,\cdots,r_n)$. Paragraph encoder takes concatenation of word representation of the start word and end word as input and returns paragraph representation. Each encoder layer is composed of two sub-layers namely a multi-head self attention layer (Section~\ref{sec:mhatt}) and a position wise fully connected feed forward neural network (Section~\ref{sec:posenc}). To be able to effectively describe these modules, we will benefit first from a description of the decoder (Section~\ref{sec:decoder}). 

\subsubsection{Decoder}
\label{sec:decoder}
The decoder stack is similar to encoder stack except that it has an additional sub layer (encoder-decoder attention layer) which learn multi-head self attention over the output of the paragraph encoder.

The output of the paragraph encoder is transformed into a set of attention vectors $K_{encdec}$ and $V_{encdec}$. Encoder-decoder attention layer of decoder takes the key $K_{encdec}$ and value $V_{encdec}$ . Decoder stack will output a float vector, we can feed this float vector to a linear followed softmax layer to get probability for generating target word.

\subsubsection{The HATT and MHATT Modules} 
\label{sec:mhatt}
Let us assume that the question decoder needs to attend to the source paragraph during the generation process.
To attend to the hierarchical paragraph representation, we replace the multi-head attention mechanism (to the source) in Transformer by introducing a new multi-head hierarchical attention module $MHATT(q^s, K^s, q^w, K^w, V^w)$ where $q^s$ is the sentence-level query vector, $q^w$ is the word level query vector, $K^s$ is the key matrix for the sentences of the paragraph, $K^w$ is the key matrix for the words of the paragraph, and $V^w$ is the value matrix fr the words of the paragraph. 

The vectors of sentence-level query $q^s$ and word-level query $q^s$  are created using non-linear transformations of the state of the decoder $\pmb{h}_{t-1}$, i.e. the input vector to the softmax function when generating the previous word $w_{t-1}$ of the question. 
The matrices for the sentence-level key $K^s$ and word-level key $K^w$ are created using the output. 
We take the input vector to the softmax function $\pmb{h}_{t-1}$, when the $t$-th word in the question is being generated.
Firstly, this module attends to paragraph sentences using their keys and the sentence query vector: 
$\pmb{a} = softmax(q_s K_s/d)$.
Here, $d$ is the dimension of the query/key vectors; the dimension of the resulting attention vector would be the number of sentences in the paragraph.  
Secondly, it computes an attention vector for the words of each sentence:
$\pmb{b}_i = softmax(q_w K_w^i/d)$. 
Here, $K^i_w$ is the key matrix for the words in the $i$-th sentences; the dimension of the resulting attention vector $\pmb{b}_i$ is the number of tokens in the $i$-th sentence. %
Lastly, the context vector is computed using the word values of their attention weights based on their sentence-level and word-level attentions: 
$$HATT(q_s, K_s, q_w, K_w, V_w) = \sum_{i=1}^{|\pmb{d}|} a_i \big( \pmb{b}_i \cdot V^w_i \big)$$
Attention in MHATT module is calculated as:
\begin{equation}
    \textit{Attention}(Q_w,V_w,K_w) = \textit{softmax}(\frac{Q_wK_w}{\sqrt{d_{k}}})V_w
\end{equation}
Where \textit{Attention}$(Q_w,V_w,K_w)$ is reformulation of scaled dot product attention of \citep{vaswani2017attention}. For multiple heads, the multihead attention $\textbf{z}= \textit{Multihead}(Q_w,K_w,V_w)$ is calculated as:
\begin{equation}
    \textit{Multihead}(Q_w,K_w,V_w) = \textit{Concat}(h_0,h_2,..,h_n)W^O
\end{equation}
where $\ h_i = \textit{Attention}(Q_wW_i^Q,K_wW_i^K,V_wW_i^V)$, $W_i^Q\in R^{d_{model}\times d_k}$, $W_i^K \in R^{d_{model}\times d_k}$ , $W_i^V \in R^{d_{model}\times d_v}$, $W^O \in R^{hd_v \times d_{model}}$, $d_k=d_v=d_{model}/h=64$.\\

\textbf{z} is fed to a position-wise fully connected feed forward neural network to obtain the final input representation.

\subsubsection{Position-wise Fully Connected Feed Forward Neural Network \label{sec:posenc}}
Output of the HATT module is passed to a fully connected feed forward neural net (FFNN) for calculating the hierarchical representation of input (\textbf{r}) as: 
$\textbf{r} = FFNN(x) = (max(0,xW_1+b1))W_2+b_2$, where 
\textbf{r} is fed as input to the next encoder layers. The final representation \textbf{r} from last layer of decoder is fed to the linear followed by softmax layer for calculating output probabilities.

\section{Experimental Setup}

\subsection{Datasets}
We performed all our experiments on the publicly available SQuAD \citep{rajpurkar2016squad} and MS MARCO~\citep{nguyen2016ms} datasets. SQuAD contains 536 Wikipedia articles and more than 100K questions posed about
the articles by crowd-workers. We split the SQuAD train set by the ratio 90\%-10\% into train and dev set and take SQuAD dev set as our test set for evaluation. We take an entire paragraph in each train/test instance as input in all our experiments. MS MARCO contains passages that are retrieved from web documents and the questions are anonimized versions of BING queries. We take a subset of MS MARCO v1.1 dataset containing questions that are answerable from atleast one paragraph. We split train set as 90\%-10\% into train (71k) and dev (8k) and take dev set as test set (9.5k). Our split is same but our dataset also contains (para, question) tuples whose answers are not a subspan of the paragraph, thus making our task more difficult.

\subsection{Evaluation metrics}
For evaluating our question generation model we report the standard metrics, {\em viz.}, BLEU~\citep{papineni2002bleu} and ROUGE-L\citep{Lin2004ROUGEAP}. We performed human evaluation to further analyze quality of questions generated by all the models. We analyzed quality of questions generated on a) syntactic correctness b) semantic correctness and c) relevance to the given paragraph. 

\subsection{Models}
We compare QG results of our hierarchical LSTM and hierarchical Transformer with their \emph{flat} counterparts. We describe our models below:

\noindent \textbf{\seq} is the attention-based sequence-to-sequence model with a BiLSTM encoder, answer encoding and an LSTM decoder.

\noindent \textbf{\hseq} is the hierarchical BiLSTM model with a BiLSTM sentence encoder, a BiLSTM paragraph encoder and an LSTM decoder conditioned on encoded answer.

\noindent \textbf{\tseq} is a Transformer-based sequence-to-sequence model with a Transformer encoder followed by a Transformer decoder conditioned on encoded answer. 

\noindent \textbf{\htseq} is the hierarchical Transformer model with a Transformer sentence encoder, a Transformer paragraph encoder followed by a Transformer decoder conditioned on answer encoded.

\begin{table*}[!htb]
    \centering
    \begin{tabular}{| l | c | c | c | c |c | c | }
    \hline
    Model& BLEU-1 & BLEU-2 & BLEU-3 & BLEU-4 &ROUGE-L \\ \hline
          
          \seq & 52.86 &29.02 & 17.06& 10.26 &38.17 \\
         \tseq & 42.07 & 22.03 & 12.33 & 7.45  & 36.77\\
         \hline
         
         \hseq  & 54.36  &\textbf{30.62} &\textbf{18.43} &\textbf{11.50}  & 38.83\\
         \htseq & \textbf{54.49} & 29.79 &17.45 &10.80 & \textbf{41.13}\\
         \hline
    \end{tabular}
    \caption{Automatic evaluation results on the SQuAD dataset. For each metric, best result is \textbf{bolded} }
    \label{tab:res_squad}
\end{table*}

\begin{table*}[!htb]
    \centering
    \begin{tabular}{| l | c | c | c | c |c | c | }
    \hline
    Model& BLEU-1 & BLEU-2 & BLEU-3 & BLEU-4 &ROUGE-L \\ \hline
          
         \seq  & 35.90 &23.52 & 15.15& 10.10 &31.05 \\
         \tseq & 24.90 & 15.32 &9.46 &6.13 & 30.76\\
         \hline

        \hseq  & \textbf{38.08}  &\textbf{25.33} &\textbf{16.48} &\textbf{11.13} &\textbf{32.82}\\
         \htseq & 31.49 & 20.05 & 12.60 & 8.68 &31.88\\
        
         \hline
    \end{tabular}
    \caption{Automatic evaluation results on the MS MARCO dataset. For each metric, best result is \textbf{bolded} }
    \label{tab:res_msmarco}
\end{table*}

\begin{table*}[!htb]
\begin{center}
\begin{tabular}{| l | c | c | c | c |c | c | }
    \hline
     \multirow{2}{*}{Model}& \multicolumn{2}{c|}{Syntax} & \multicolumn{2}{c|}{Semantics} & \multicolumn{2}{c|}{Relevance} \\ \cline{2-7}
            & Score & Kappa & Score & Kappa & Score & Kappa \\
    \hline
     \seq &86 &0.57 & 79.33 &0.61 &70.66 & 0.56  \\
     \tseq &86 &0.66 & 84 &0.62 &50 & 0.60  \\
     \hline
     \hseq &80 &0.49 & 73.33 &0.54 &\textbf{81.33} & 0.64  \\
      \htseq &\textbf{90} &0.62 & \textbf{85.33} &0.65 &68 & 0.56  \\
\hline
\end{tabular}
  
\end{center}
\caption{Human evaluation results (column ``Score'') as well as inter-rater agreement (column ``Kappa'') for each model on the SQuAD test set. The scores are between 0-100, 0 being the worst and 100 being the best. Best results for each metric (column) are \textbf{bolded}. The three evaluation criteria are: (1) syntactically correct ({Syntax}), (2) semantically correct ({Semantics}), and (3) relevant to the text ({Relevance}).}
\label{heresults_squad}
\end{table*}

\begin{table*}[!htb]
\begin{center}
\begin{tabular}{| l | c | c | c | c |c | c | }
    \hline
     \multirow{2}{*}{Model}& \multicolumn{2}{c|}{Syntax} & \multicolumn{2}{c|}{Semantics} & \multicolumn{2}{c|}{Relevance} \\ \cline{2-7}
            & Score & Kappa & Score & Kappa & Score & Kappa \\
    \hline
     \seq &83.33 &0.68 & 69.33 & 0.65 &38.66 & 0.52 \\
     \tseq &80.66 &0.73 & 73.33 &0.55 &35.33 & 0.47  \\
     \hline
     \hseq &85.33 &0.73 & 70.66 &0.68 &\textbf{51.33} & 0.60  \\
      \htseq &\textbf{86} &0.87 & \textbf{73.33} &0.65 &32.66 & 0.60  \\
\hline
  \end{tabular}
  
\end{center}
\caption{Human evaluation results (column ``Score'') as well as inter-rater agreement (column ``Kappa'') for each model on the MS MARCO test set. The scores are between 0-100, 0 being the worst and 100 being the best. Best results for each metric (column) are \textbf{bolded}. The three evaluation criteria are: (1) syntactically correct ({Syntax}), (2) semantically correct ({Semantics}), and (3) relevant to the text ({Relevance}).}
\label{heresults_msmarco}

\end{table*}

\section{Results and discussion}
In Table \ref{tab:res_squad} and Table \ref{tab:res_msmarco} we present automatic evaluation results of all models on SQuAD and MS MARCO datasets respectively. We present human evaluation results in Table \ref{heresults_squad} and  Table \ref{heresults_msmarco} respectively. 

A number of interesting observations can be made from automatic evaluation results in Table \ref{tab:res_squad} and Table \ref{tab:res_msmarco}: 

\begin{itemize}
\item Overall, the hierarchical BiLSTM model \hseq shows the best performance, achieving best result on BLEU2--BLEU4 metrics on both SQuAD dataset, whereas the hierarchical Transformer model \tseq\ performs best on BLEU1 and ROUGE-L on the SQuAD dataset.

\item Compared to the flat LSTM and Transformer models, their respective hierarchical counterparts always perform better on both the SQuAD and MS MARCO datasets.


\item On the MS MARCO dataset, we observe the best consistent performance using the hierarchical BiLSTM models on all automatic evaluation metrics.

\item On the MS MARCO dataset, the two LSTM-based models outperform the two Transformer-based models. 

\end{itemize}

Interestingly, human evaluation results, as tabulated in Table \ref{heresults_squad} and  Table \ref{heresults_msmarco}, demonstrate that the hierarchical Transformer model \tseq\ outperforms all the other models on both datasets in both syntactic and semantic correctness. However, the hierarchical BiLSTM model \hseq\ achieves best, and significantly better, relevance scores on both datasets. 

From the evaluation results, we can see that our proposed hierarchical models demonstrate benefits over their respective flat counterparts in a significant way. Thus, for paragraph-level question generation, the hierarchical representation of paragraphs is a worthy pursuit. Moreover, the Transformer architecture shows great potential over the more traditional RNN models such as BiLSTM as shown in human evaluation. Thus the continued investigation of hierarchical Transformer is a promising research avenue.  In the Appendix, in Section B, we present several examples that illustrate the effectiveness of our Hierarchical models. In Section C of the appendix, we present some failure cases of our model, along with plausible explanations.

\section{Conclusion}
We proposed two hierarchical models for the challenging task of question generation from paragraphs, one of which is based on a hierarchical BiLSTM model and the other is a novel hierarchical Transformer architecture. 
We perform extensive experimental evaluation on the SQuAD and MS MARCO datasets using standard metrics. Results demonstrate the hierarchical representations to be overall much more effective than their flat counterparts. The hierarchical models for both Transformer and BiLSTM clearly outperforms their flat counterparts on all metrics in almost all cases. Further, our experimental results validate that hierarchical selective attention benefits the hierarchical BiLSTM model. Qualitatively, our hierarchical models also exhibit better capability of generating fluent and relevant questions. 

\bibliography{anthology,acl2020}

\begin{thebibliography}{20}
\expandafter\ifx\csname natexlab\endcsname\relax\def\natexlab#1{#1}\fi

\bibitem[{Ali et~al.(2010)Ali, Chali, and Hasan}]{ali2010automation}
Husam Ali, Yllias Chali, and Sadid~A Hasan. 2010.
\newblock Automation of question generation from sentences.
\newblock In \emph{Proceedings of QG2010: The Third Workshop on Question
  Generation}, pages 58--67.

\bibitem[{Du and Cardie(2018)}]{cardie2018harvesting}
Xinya Du and Claire Cardie. 2018.
\newblock Harvesting paragraph-level question-answer pairs from {Wikipedia}.
\newblock In \emph{ACL (1)}, pages 1907--1917.

\bibitem[{Du et~al.(2017)Du, Shao, and Cardie}]{du2017learning}
Xinya Du, Junru Shao, and Claire Cardie. 2017.
\newblock Learning to ask: Neural question generation for reading
  comprehension.
\newblock In \emph{Proceedings of the 55th Annual Meeting of the Association
  for Computational Linguistics (Volume 1: Long Papers)}, volume~1, pages
  1342--1352.

\bibitem[{Fan et~al.(2018)Fan, Wei, Li, Lan, and Huang}]{fan2018question}
Zhihao Fan, Zhongyu Wei, Piji Li, Yanyan Lan, and Xuanjing Huang. 2018.
\newblock A question type driven framework to diversify visual question
  generation.
\newblock In \emph{IJCAI}, pages 4048--4054.

\bibitem[{Goldberg(2019)}]{goldberg2019assessing}
Yoav Goldberg. 2019.
\newblock Assessing bert's syntactic abilities.
\newblock \emph{CoRR}, abs/1901.05287.

\bibitem[{Heilman(2011)}]{heilman2011automatic}
Michael Heilman. 2011.
\newblock \emph{Automatic factual question generation from text}.
\newblock Ph.D. thesis, Carnegie Mellon University.

\bibitem[{Heilman and Smith(2010)}]{heilman2010good}
Michael Heilman and Noah~A Smith. 2010.
\newblock Good question! statistical ranking for question generation.
\newblock In \emph{Human Language Technologies: The 2010 Annual Conference of
  the North American Chapter of the Association for Computational Linguistics},
  pages 609--617. Association for Computational Linguistics.

\bibitem[{Kumar et~al.(2018)Kumar, Boorla, Meena, Ramakrishnan, and
  Li}]{kumar2018automating}
Vishwajeet Kumar, Kireeti Boorla, Yogesh Meena, Ganesh Ramakrishnan, and
  Yuan-Fang Li. 2018.
\newblock Automating reading comprehension by generating question and answer
  pairs.
\newblock In \emph{Pacific-Asia Conference on Knowledge Discovery and Data
  Mining}, pages 335--348. Springer.

\bibitem[{Kumar et~al.(2019)Kumar, Ramakrishnan, and Li}]{kumar2018framework}
Vishwajeet Kumar, Ganesh Ramakrishnan, and Yuan-Fang Li. 2019.
\newblock Putting the horse before the cart: A generator-evaluator framework
  for question generation from text.
\newblock In \emph{The SIGNLL Conference on Computational Natural Language
  Learning (CoNLL 2019)}.

\bibitem[{Li et~al.(2018)Li, Duan, Zhou, Chu, Ouyang, Wang, and
  Zhou}]{li2018visual}
Yikang Li, Nan Duan, Bolei Zhou, Xiao Chu, Wanli Ouyang, Xiaogang Wang, and
  Ming Zhou. 2018.
\newblock Visual question generation as dual task of visual question answering.
\newblock In \emph{Proceedings of the IEEE Conference on Computer Vision and
  Pattern Recognition}, pages 6116--6124.

\bibitem[{Lin(2004)}]{Lin2004ROUGEAP}
Chin-Yew Lin. 2004.
\newblock Rouge: A package for automatic evaluation of summaries.
\newblock In \emph{ACL 2004}.

\bibitem[{Luong et~al.(2015)Luong, Pham, and Manning}]{Luong2015EffectiveAT}
Thang Luong, Hieu~Quang Pham, and Christopher~D. Manning. 2015.
\newblock Effective approaches to attention-based neural machine translation.
\newblock In \emph{EMNLP}.

\bibitem[{Nguyen et~al.(2016)Nguyen, Rosenberg, Song, Gao, Tiwary, Majumder,
  and Deng}]{nguyen2016ms}
Tri Nguyen, Mir Rosenberg, Xia Song, Jianfeng Gao, Saurabh Tiwary, Rangan
  Majumder, and Li~Deng. 2016.
\newblock Ms marco: A human generated machine reading comprehension dataset.
\newblock \emph{arXiv preprint arXiv:1611.09268}.

\bibitem[{Papineni et~al.(2002)Papineni, Roukos, Ward, and
  Zhu}]{papineni2002bleu}
Kishore Papineni, Salim Roukos, Todd Ward, and Wei-Jing Zhu. 2002.
\newblock {BLEU}: a method for automatic evaluation of machine translation.
\newblock In \emph{Proceedings of the 40th annual meeting on association for
  computational linguistics}, pages 311--318. Association for Computational
  Linguistics.

\bibitem[{Pennington et~al.(2014)Pennington, Socher, and
  Manning}]{pennington-etal-2014-glove}
Jeffrey Pennington, Richard Socher, and Christopher Manning. 2014.
\newblock \href {https://doi.org/10.3115/v1/D14-1162} {{G}love: Global vectors
  for word representation}.
\newblock In \emph{Proceedings of the 2014 Conference on Empirical Methods in
  Natural Language Processing ({EMNLP})}, pages 1532--1543, Doha, Qatar.
  Association for Computational Linguistics.

\bibitem[{Rajpurkar et~al.(2016)Rajpurkar, Zhang, Lopyrev, and
  Liang}]{rajpurkar2016squad}
Pranav Rajpurkar, Jian Zhang, Konstantin Lopyrev, and Percy Liang. 2016.
\newblock {SQuAD}: 100,000+ questions for machine comprehension of text.
\newblock In \emph{Proceedings of the 2016 Conference on Empirical Methods in
  Natural Language Processing}, pages 2383--2392.

\bibitem[{Song et~al.(2018)Song, Wang, Hamza, Zhang, and
  Gildea}]{song2018leveraging}
Linfeng Song, Zhiguo Wang, Wael Hamza, Yue Zhang, and Daniel Gildea. 2018.
\newblock Leveraging context information for natural question generation.
\newblock In \emph{Proceedings of the 2018 Conference of the North American
  Chapter of the Association for Computational Linguistics: Human Language
  Technologies, Volume 2 (Short Papers)}, volume~2, pages 569--574.

\bibitem[{Tran et~al.(2018)Tran, Bisazza, and Monz}]{tran2018importance}
Ke~Tran, Arianna Bisazza, and Christof Monz. 2018.
\newblock The importance of being recurrent for modeling hierarchical
  structure.
\newblock \emph{arXiv preprint arXiv:1803.03585}.

\bibitem[{Vaswani et~al.(2017)Vaswani, Shazeer, Parmar, Uszkoreit, Jones,
  Gomez, Kaiser, and Polosukhin}]{vaswani2017attention}
Ashish Vaswani, Noam Shazeer, Niki Parmar, Jakob Uszkoreit, Llion Jones,
  Aidan~N Gomez, {\L}ukasz Kaiser, and Illia Polosukhin. 2017.
\newblock Attention is all you need.
\newblock In \emph{Advances in neural information processing systems}, pages
  5998--6008.

\bibitem[{Zhao et~al.(2018)Zhao, Ni, Ding, and Ke}]{zhao2018paragraph}
Yao Zhao, Xiaochuan Ni, Yuanyuan Ding, and Qifa Ke. 2018.
\newblock Paragraph-level neural question generation with maxout pointer and
  gated self-attention networks.
\newblock In \emph{Proceedings of the 2018 Conference on Empirical Methods in
  Natural Language Processing}, pages 3901--3910.

\end{thebibliography}
\bibliographystyle{acl_natbib}

\end{document}